\documentclass[letterpaper]{article} 
\usepackage{aaai25}  
\usepackage{times}  
\usepackage{helvet}  
\usepackage{courier}  
\usepackage[hyphens]{url}  
\usepackage{graphicx} 
\usepackage{natbib}  
\usepackage{caption} 
\usepackage{algorithm}
\usepackage{algorithmic}

\usepackage{tikz}

\newcommand*\circled[1]{%
    \tikz[baseline=(char.base)]{
        \node[shape=circle,draw,inner sep=0pt] (char) {#1};
    }%
}

\usepackage{xcolor}

\usepackage{newfloat}
\usepackage{listings}
\DeclareCaptionStyle{ruled}{labelfont=normalfont,labelsep=colon,strut=off} 
\floatstyle{ruled}
\newfloat{listing}{tb}{lst}{}
\floatname{listing}{Listing}
\title{Towards Personalized Explanations for Health Simulations:\\A Mixed-Methods Framework for Stakeholder-Centric Summarization}
\author{
    Philippe J. Giabbanelli\textsuperscript{\rm 1} and 	
Ameeta Agrawal\textsuperscript{\rm 2}
}
\affiliations{
    \textsuperscript{\rm 1}Virginia Modeling, Analysis, and Simulation Center (VMASC), Old Dominion University\\
    1030 University Blvd, Suffolk, VA 23435, USA\\
    pgiabban@odu.edu

    \textsuperscript{\rm 2}Department of Computer Science, Portland State University\\
    1900 SW 4th Ave, Portland, OR 97201, USA\\
    ameeta@pdx.edu
}

\begin{document}

\maketitle

\begin{abstract}
Modeling \& Simulation (M\&S) approaches such as agent-based models hold significant potential to support decision-making activities in health, with recent examples including the adoption of vaccines, and a vast literature on healthy eating behaviors and physical activity behaviors. These models are potentially usable by different stakeholder groups, as they support policy-makers to estimate the consequences of potential interventions and they can guide individuals in making healthy choices in complex environments. However, this potential may not be fully realized because of the models' complexity, which makes them inaccessible to the stakeholders who could benefit the most. While Large Language Models (LLMs) can translate simulation outputs and the design of models into text, current approaches typically rely on one-size-fits-all summaries that fail to reflect the varied informational needs and stylistic preferences of clinicians, policymakers, patients, caregivers, and health advocates. This limitation stems from a fundamental gap: we lack a systematic understanding of what these stakeholders need from explanations and how to tailor them accordingly. To address this gap, we present a step-by-step framework to identify stakeholder needs and guide LLMs in generating tailored explanations of health simulations. Our procedure uses a mixed-methods design by first eliciting the explanation needs and stylistic preferences of diverse health stakeholders, then optimizing the ability of LLMs to generate tailored outputs (e.g., via controllable attribute tuning), and then evaluating through a comprehensive range of metrics to further improve the tailored generation of summaries.
\end{abstract}

%

\section{Introduction}
Simulation models have been developed for many health application scenarios, such as finding the right treatment for depression given a \textit{patient}'s characteristics~\cite{wittenborn2022exploring}, supporting \textit{policymakers} in identifying efficient campaigns to promote healthy eating and reduce hypertension~\cite{khademi2018agent}, or optimizing the layout of a hospital~\cite{dos2025enhancing}. These models can be complex: for example, the hypertension model accounts for the structure of social networks and the diffusion of social norms, as well as individual taste preferences and the eventual impact of food consumption patterns onto health outcomes. This complexity may create barriers to participation in the modeling process for non-modelers, particularly compounded with recurring issues related to model communication and low levels of traceability~\cite{belfrage2024simulating}. These challenges may partly explain recent evidence from the literature on human-centered computing in which participants were primarily engaged in the early stage of model building but insufficiently as development progressed~\cite{manellanga2024participatory}. We consider that such lack of participation is a missed opportunity for validation (e.g., community members could compare the simulated journeys of agents to their own experiences) and can reduce buy-in. \citet{ahrweiler2019co} emphasized the importance of trust for buy-in: ``the first and most important is that the clients want to understand the model[:] to trust results means to trust the process that produced them.''  

Given \textit{(i)} the significant efforts devoted to building models and their demonstrated relevance for decision-making in the context of health and \textit{(ii)} the challenges of engaging participants into the process to ensure the accuracy of findings and their translation into practices, many studies have proposed and evaluated means of engaging participants into the modeling process. Methods can vary based on criteria such as the targeted degree of participation, which ranges from providing information to co-deciding~\cite{ferrand2024engineering}. \textit{We focus on the dissemination of results}, which is located as a low degree of participation. 

\begin{figure*}[!t]
\centering
\includegraphics[width=0.9\textwidth]{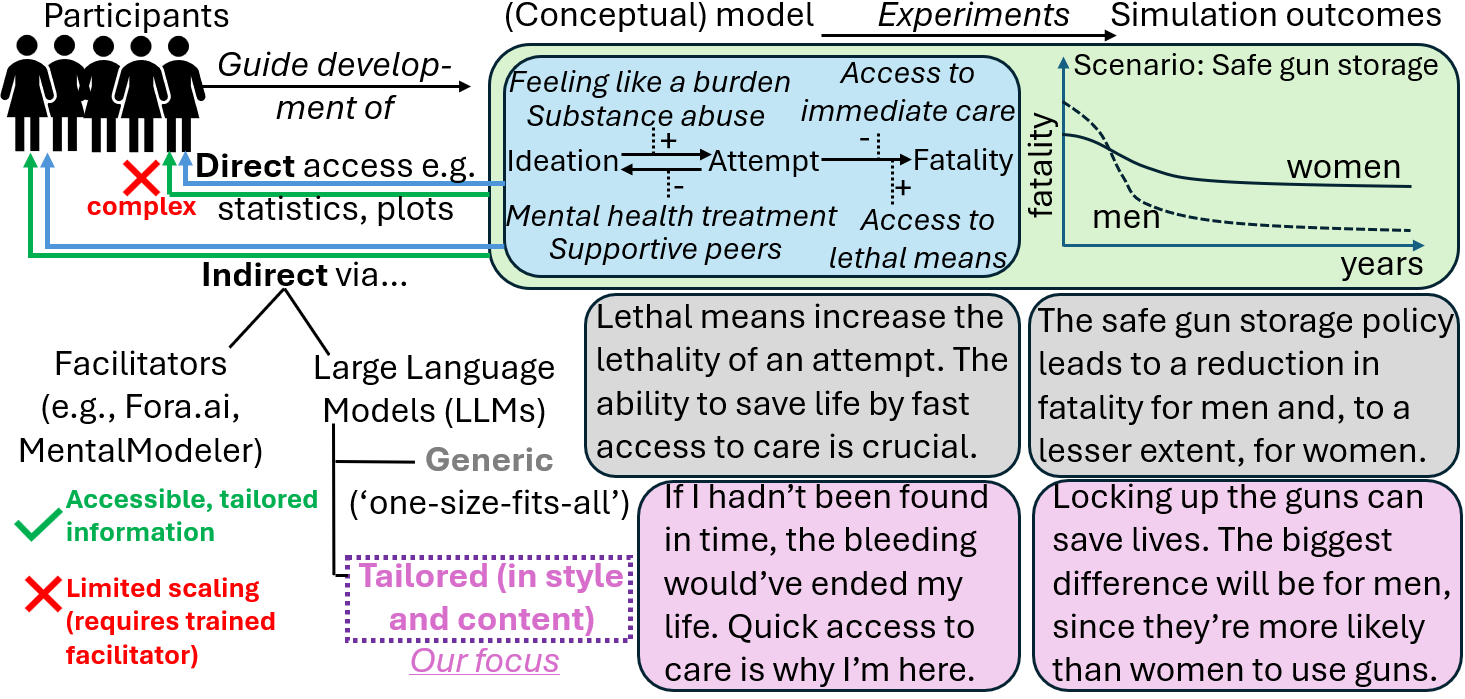} 
\caption{A model consists of elements and interrelationships from the problem domain, exemplified here as suicide prevention. The model can be programmed and then used for computational experiments, leading into simulation outcomes that can be differentiated by target populations. Our focus is on explaining models and simulations in a tailored manner (here for a lay audience) instead of the dominant and generic model-to-text translation via LLMs. This is more scalable than human facilitation.}
\label{fig:overview}
\end{figure*}

As empirical studies on information dissemination and simulations have demonstrated, participants did not ``want to look at a multitude of
tables and scan through simulation results for interesting parameters; nor did they expect to watch the running model producing its results''~\cite{ahrweiler2019co} (Fig.~\ref{fig:overview}). Researchers have co-developed interactive visualization environments with practitioners to explore models~\cite{kammler2023towards}, as exemplified by our work on policy-making and obesity in which policymakers could find key constructs and interrelationships in the model~\cite{giabbanelli2018navigating}. However, ensuing usability studies showed that such platforms had a problematic learning curve or took too long to perform simple tasks~\cite{giabbanelli2023human}. As a result, computational solutions to promote engagement with simulation models (particularly to explain a system) belong to two broad categories. 

First, there are platforms that are intended to \textit{facilitate} engagement, typically within a workshop setting with a trained facilitator. {\ttfamily Fora.ai} allows participants to express the outcomes that they wish to see from the simulations (i.e., their `concerns') and the results are primarily communicated on a map, given the platform's emphasis on environmental problems such as flooding~\cite{zellner2025enhancing}. {\ttfamily MentalModeler} supports participants in externalizing their mental models as Fuzzy Cognitive Map (visualized as node-and-link diagrams) and to see simulation outcomes in easily interpretable visuals showing which factors would increase or decrease~\cite{gray2013mental}. This tool has been used for over a decade, and the participatory technique of Fuzzy Cognitive Mapping has been abundantly used in health~\cite{sarmiento2024fuzzy} and medicine~\cite{apostolopoulos2024fuzzy}. However, trained facilitators are not always available, so this approach is more appropriate for workshop settings where a small number of participants can be guided to learn new approaches than for large-scale deployment in which participants should independently access information. This scaling limitation echoes concerns on technology exclusion about who will be granted the benefits of new platforms for decision-making~\cite{ahrweiler2025inclusive}. 

In contrast, the second approach uses technology to explain simulation results in familiar formats that people can interpret \textit{independently}, which is the focus of this paper. This approach does not have facilitators and does not currently solicit feedback, unlike the platforms mentioned above. Rather, the focus is solely on information dissemination. Large Language Models (LLMs) are prominent in this space, as they can transform data into textual summaries (Fig.~\ref{fig:overview}, grey boxes). LLMs and their uses have evolved rapidly to explain health simulations, starting the first work in 2022 that explained obesity and suicide models via GPT-3~\cite{shrestha2022automatically}. The focus was on decomposing large models so they could be `fed' to GPT one piece at a time, leading to a focus on sentence-level generation. This early prototype suffered from both the fluency issues of GPT-3 (e.g., typos, grammatical mistakes) and the limitations of lossy decomposition algorithms, which removed some of the model's aspect. Progress on LLMs resulted in high fluency, coverage, and faithfulness scores, while advances in model-to-text generation produced paragraphs (rather than bags of sentences) with coherent themes and transitions~\cite{gandee2024combining}. As technology matured, the scope was broadened from explaining the \textit{structure} of a model (e.g., how do risk and preventive factors \textit{generally} explain the transition from suicide ideation to attempt) to covering the \textit{dynamics} of a simulation (e.g., how did a \textit{simulated individual} receive mental health treatment after a non-lethal suicide attempt). For example, the latest framework focused on empathy to explain the simulated life of an individual (e.g., who they are, where they lived, how they dealt with stress), which provided a sense of immersion to readers based on a human study~\cite{giabbanelli2025promoting}. 

The emphasis has been on technical advancements in LLMs and in systems built around them, including prompts, retrieval augmented generation (RAG), and mixed-methods approaches. However, we argue that this progress has not yet resulted in achieving the fundamental mission of LLM-generated explanations for modeling and simulation in health: how do we communicate information that is understandable and actionable to each stakeholder? In particular, all studies on model-to-text translation via LLMs have taken a one-size-fits-all approach, but \textit{communication should be tailored to address the varied needs of different stakeholders}. Our vision is to move from one-size-fits-all AI-driven solutions to tailored communication that responds to individual needs, echoing some of the motivations found in the emerging field of participatory AI~\cite{ahrweiler2025inclusive}.

To appreciate why different people need different information, consider our introductory example in which a model serves to optimize the layout of a hospital. Hospital administrators need to estimate how the proposed layout would affect throughput and staffing needs, medical practitioners require operational logic (e.g., will the proximity of ICU to ER help to keep the average delay low?), patients and caregivers need to navigate the space, and public health officials need a layout that supports emergency preparedness requirements. In addition to different \textit{content} needs, we also expect differences in \textit{style}. For example, hospital administrators may need executive summaries with bullet points to support decision-making activities and a business-oriented language, whereas medical practitioners may favor terminology aligned with clinical workflows, and patients may appreciate an accessible and empathetic language.

\textbf{The main contribution} of this paper is a vision and framework for eliciting, incorporating, and evaluating the style and content needs of diverse stakeholders to obtain tailored information on health models and simulations.

The remainder of this paper is organized as follows. We provide a succinct background on the importance and the technical feasibility of generating tailored summaries of a model for different users and applications. Next, we propose a framework that identifies information needs (styles and content) from different stakeholder groups, then iteratively optimizes the alignment between LLM-generated summaries and the participants' needs. Given the growing importance of participatory AI, this paper ends with a brief discussion on how various forms of participation (from workshops to broader engagements) can contribute to design and evaluation of LLM-generated explanations.

\section{Background}
\subsection{The Importance and Challenges of Transparency in Models and Simulations for Health}
Models and simulations can support collaborative searches for solutions when complex problems are characterized by the needs for trade-offs, particularly monetary costs and differential impacts across populations. In addition, unintended consequences may be more readily identified and mitigated when groups of stakeholders examine the implications of a decision and how its modeling assumptions would perform in practice. For example, spatial analyses of obesity patterns show a close relationship between what people are exposed to (e.g., fast-food outlets near their home or workplace) and their weight~\cite{patterson2025combined}. Public health policies on childhood obesity, such as restricting the ability of new fast-food outlets to open near schools, are acceptable to young people~\cite{savory2025does} and face little opposition, especially where fast-foods are already common.~\cite{keeble2024public}. Models and simulations can help to design zoning policies, such as finding the minimal distance between fast-food outlets or with respect to schools, to achieve a target reduction in consumption over a simulated period~\cite{baniukiewicz2018capturing}. However, planners note that businesses can change the business category to avoid regulation or open within other locations~\cite{hassan2024adoption}, so unintended consequences include a \textit{displacement} of outlets (e.g. to deprived areas) or a transformation of the food landscape that does not support healthy eating. If business owners, young people, planners, and obesity researchers could examine simulated zoning policies, they may note such consequences and either propose revisions to the model or identify pilot areas that are sufficiently constrained for the model to operate as intended.

Stakeholders' shared understanding of the problem and the novelty, concreteness, and richness of proposed solutions evolve alongside a model's degree of realism, but up to a certain point. While more realistic models can yield more detailed estimates or investigate complex tradeoffs, they can also overwhelm users, preventing them from acting on the insights they derived from the model, particularly within social contexts that exhibit strong power dynamics and favor prediction~\cite{zellner2022finding}. Simpler models may be better understood, resulting in faster analyses and improved implementations~\cite{brooks1996choosing}, even if they do not provide the most accurate support for decision-making activities. There is thus evidence of the importance to balance representational fidelity and \textit{end-user intelligibility}. 

While there is abundant advice from researchers on using policy models for decision-making~\cite{ghaffarzadegan2011small,luna2006anatomy}, actual evidence of policymakers using simulation models to inform policy choices is limited~\cite{ahrweiler2019co}, e.g. to unique health challenges such as pandemic planning~\cite{janssen2018innovating,haddad2020crisis}. As the lack of end-user intelligibility is a key challenge, improving model transparency may enhance the likelihood of producing models that stakeholders can use. 

A model may be better understood once simplified, either by removing inappropriate complexity (e.g., redundant variables) or by manually performing subtle alterations (e.g., exclude infrequent events, replacing feedback loops by constant) and checking that results between a simplified model align with its original version~\cite{robinson2024assumptions,van2005strategy}. However, there are cautionary tales in the case of empirically-grounded individual level models such as Agent-Based Models (ABMs), which are of particular interest for health simulations. As noted by~\citet{sun2016simple}, ABMs are generally created to provide predictions based on the specific needs of users at specific places, so oversimplification can decrease a model's usefulness, validity, and credibility~\cite{van2017approaches}. For example, modelers have expressed concerns that simple models are usually achieved by ignoring spatial heterogeneity and individual variability~\cite{van2017approaches}, which would overlook how policies work differently across populations~\cite{yu2024disparities} and places (e.g., access to care, exposure to risk factors). In addition, when critical aspects are missed due to oversimplification, users may resort to their motivational biases to fill-in for missing information, for example by perceiving events as more or less likely depending on whether they are desirable~\cite{montibeller2018behavioral}. This can result in choosing a policy option based on biases rather than scientific models.

\subsection{Generating Text to Address the Need for Simulation Transparency}
Transparency does not only depend on the model: it also depends on the users. Some aspects of a model may be well-understood by one set of users because it relates to their lived experiences, while other aspects may quickly create a cognitive overload. As summarized by~\citet{robinson2024assumptions}, ``a user's comprehension of [a model is] subjective, dependent on the perceptions of the model users, and their knowledge and experience''. Communicating about models thus requires an assessment of the target users, since their different mental models, expectations, and experiences bring `subjective and active elements' into the modeling exercise~\cite{hamalainen2013importance}.

When simulation results are communicated to modelers, summary statistics and data visualizations play an important role~\cite{st2023survey}. For instance, our simulation platform for suicide prevention (developed with the CDC) provides health outcomes (suicide ideation, attempt, fatality) change over time across gender or race and ethnicity~\cite{huddleston2022design}. Users can examine \textit{why} these outcomes happened by tracking the fraction of agents with specific factors such as being bullied, hopeless, or a victim of physical or sexual abuse. Results are provided as interactive visualizations as well as raw data that can be exported for statistical analyses or parsed by a screen reader for users with visual impairments. However, data visualizations and statistics are not suitable for every audience and/or they do not easily integrate in every workflow. There is thus significant research interest in complementing or replacing these artifacts by using LLMs to translate simulations into textual explanations~\cite{Fedeli,dolha2024generative,fahland2024well}.

Recent works have considered that four parts of a simulation could be explained as text: functionalities (domain-specific knowledge about the purpose and capabilities of the simulation), validity (metrics to evaluate in which situations and to which extent the results can be trusted), architecture (what inputs parameters can be modified and how do they affect observable outputs), and operations (libraries and software components involved during execution)~\cite{Fedeli}. So far, \textit{no study has examined which aspects of a simulation model should be conveyed to a given audience, in which style, or how to measure the reactions of an audience to textual summaries of simulations}.

\subsection{Tailoring Text Generation by LLMs for Different Users and Applications}

Models can involve many parameters, rules, datasets for calibration and validation, and simulation scenarios. As explained above, detailing the minute aspects of every component to every user would not be productive. Rather, we need to \textit{summarize} the aspects of a model that matter to an individual (content) in a manner that supports their decision-making activities (style). LLMs are now widely used to generate summaries~\cite{olabisi-agrawal-2024-understanding,zhang2025systematic}, with several reviews detailing applications to clinical settings~\cite{busch2025current,bednarczyk2025scientific,shool2025systematic,bedi2025testing}. Concerns about using LLMs in healthcare often focus on hallucinations, biased or stereotyped outputs~\cite{omar2025evaluating}, and inconsistent reasoning across languages~\cite{schlicht2025llms} or multiple uses of the same prompt. These risks are especially important to monitor when patient safety is at stake, and robust oversight during deployment is essential. However, demanding absolute perfection from LLMs may be counterproductive. Rejecting systems for occasional imperfections can limit access to information (e.g., end-users cannot interpret simulation results without assistance) or force reliance on human intermediaries, such as modelers, who are neither perfectly accurate nor always available. Notably, a recent study found that physicians preferred LLM-generated answers over those from other physicians on eight of nine clinical axes~\cite{singhal2025toward}. In another study, physicians found 81\% of LLM-generated summaries equivalent or superior to summaries from medical experts on completeness, correctness, and conciseness~\cite{van2024adapted}.

Given that LLMs can produce good summaries in health, the next step is to tailor these summaries to the needs and intents of individual users. For example, a generic description of scientific simulations may cover both the model and the domain, thus non-modelers would struggle with domain-specific terminology while subject matter experts may struggle with simulation techniques – as a result neither audience is satisfied. Consequently, \textit{customizable summarization} is essential to accommodate individual user preferences~\cite{bagheri-nezhad-etal-2025-fair,wang2025recent,kirstein2025cads}. Adapting to a user's needs relies on controllable attributes~\cite{urlana2023controllable}, such as summary length (to produce executive reports), writing style (to match the desired tone or message), information coverage (to include essential content), content diversity (increases the variety of topics covered), and topic control (ensures clarity by focusing on specific themes in research papers or reports). For instance, style-controlled summaries can maintain a consistent tone across different communication channels, while topic-controlled summaries enhance coherence by emphasizing specific areas of interest. Despite the technical feasibility of creating tailored summaries, this has \textit{never been explored in the context of facilitating access to scientific simulations since the needs of each user group have not been assessed}. 

\begin{figure*}[!t]
\centering
\includegraphics[width=1\textwidth]{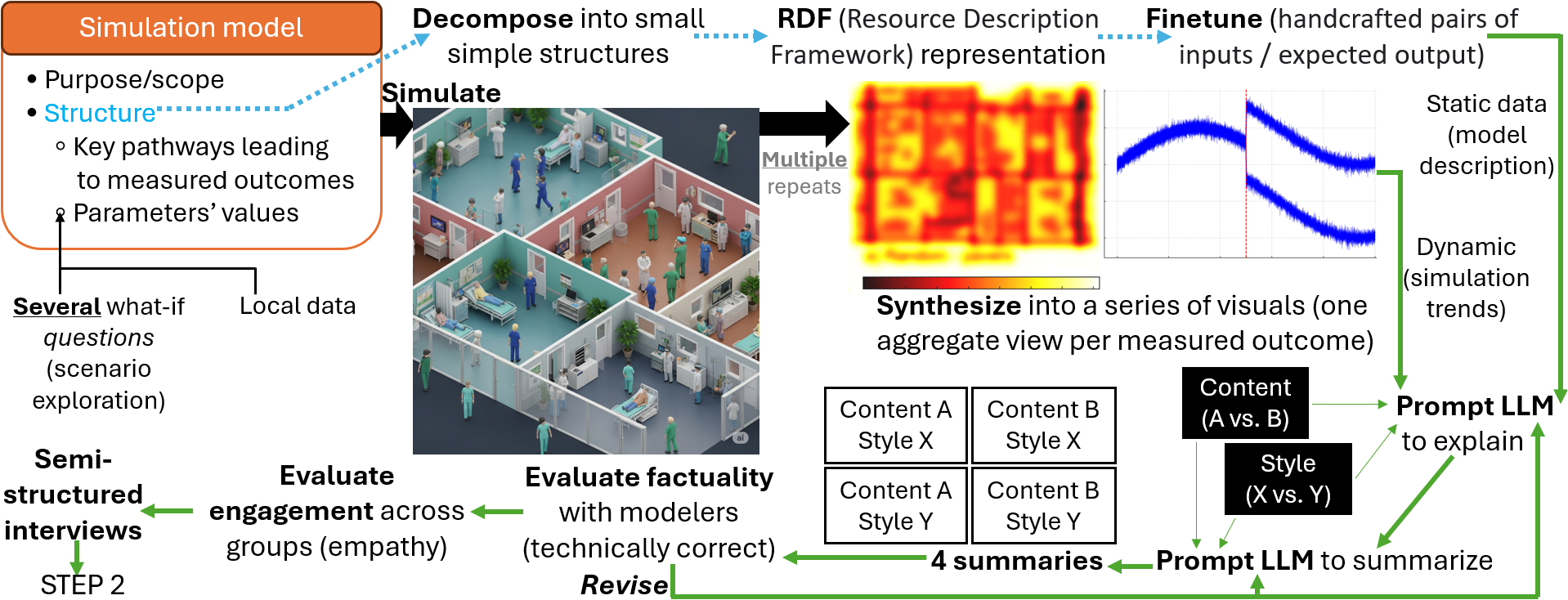} 
\caption{A simulation model has (1) a static structure, which can be decomposed and transformed into text using existing solutions \protect\cite{gandee2024combining}, and (2) dynamic datasets (e.g., the states of simulated agents over time) created in response to simulation scenarios. Static and dynamic elements are turned into summaries (as a starting point) and evaluated.}
\label{fig:Step2}
\end{figure*}

\section{Proposed Framework}
Our proposal has two broad steps, detailed in the following two subsections and summarized in Figures~\ref{fig:Step2} and~\ref{fig:Step3}. In the first step, we create technically correct summaries of a model's structure and simulation outcomes (as assessed by modelers) and we use them to measure the reactions of participants across groups to summaries. Given these needs, the second step generates several summaries to cover different potential preferences in content and style, and we re-assess them with participants for iterative feedback.

\subsection{Step 1: Identify Information Needs and Preferred Styles From Different Stakeholder Groups}

A simulation model has two parts: a \textit{static part that describes the model} (e.g., purpose, scope of what is included/excluded, main parameters and how they drive outputs) and \textit{a dynamic dataset generated through simulations}. Translating the static part proceeds through three sub-steps. 

\circled{1} First, we split the full specification of the model into smaller parts that collectively contain the information (i.e., decomposition facilitates information processing but does not lose information). Available algorithms include `RDF Walks'~\cite{mussa2024towards} and our hierarchical approach~\cite{gandee2024combining}. Breaking down complex models into sufficiently small sizes helps LLMs such as GPT in generating well-formed sentences~\cite{shrestha2022automatically}, as measured by standard text quality scores in Natural Language Processing (e.g. ROUGE, BLEU). Note that the order produced by this \textit{linearization} process (taking a graph and producing a sequential list) affects the LLM's performance, so the ordering can be optimized to preserve attribution and reduce hallucination. In addition, a hierarchical decomposition can help to structure the summary. Such decomposition shouldn't just be a technical inventory of model components, as it would be difficult to explain agents separately from physical spaces or processes. Rather, a well-formed decomposition should mirror a logical narrative that an explanation can follow, so the reader understands the model in a natural flow. For instance, a disease-progression perspective provides a useful organizing principle because it aligns with how both domain experts and lay readers think about the system: what happens first (who is in the population, what's their health status, how do they become exposed), what happens next (how an infection progresses over time), and how different processes connect (testing and diagnosis, isolation, treatment).

\circled{2} Second, we represent each part in a structured manner. The smaller structures of a model consist of triplets, where one element has a relation with another element. The triplets can be represented in different ways, such as ``head A | relation X | tail B'' or via tags as $<$head$>$A$<$/head$> <$relate$>$X$<$/relate$> <$tail$>$B$<$/tail$>$. Recent works showed that the choice of representation does have an impact with current LLMs, but it was not significant~\cite{he2025evaluating}. We thus use RDF as an example in this section, since it has been studied in several other studies on LLMs for medical applications~\cite{mavridis2025large}, but developers may use another representation.

\circled{3} Third, some health professionals use a web portal to access some LLM, type a specialized medical prompt, and find the answer unsatisfactory~\cite{gravel2023learning,ponzo2024chatgpt} -- as would be expected in using a general purpose LLM. We emphasize that LLMs should not be expected to serve as oracles equipped with universal expertise: rather, the onus is on users to correctly employ LLMs as a component in an engineered pipeline. In particular, we must ensure domain adaptation through fine-tuning and/or retrieval augmented generation with authoritative medical sources. In our context, fine-tuning could be onerous by manually crafting a large number of pairs of $<$sample model input, expected textual output$>$. In an experimental study on two health cases (suicide and obesity), we explored the response curve between investing in more fine-tuning and the quality of the output~\cite{giabbanelli2024narrating}. We found that a handful of examples (few-shot learning) were sufficient for saturation, but they are also \textit{necessary} as performances on zero-shot learning were inadequate. 

The three steps above transform the structure of a model into text by feeding the entire structure to the LLM in small chunks (`bite size'). As we turn our attention to the translation of simulations to text, we face a different problem of \textit{volume}. A simulation may involve many virtual agents, whose activities span a long time, and many simulation runs may be needed to obtain a distribution of results (e.g., characterize health outcomes with a 95\% confidence interval). Consequently, we cannot `feed' the complete simulation results to the LLM in small increments in the same way as we decomposed the whole structure. A simple approach is to perform a statistical analysis and provide it as input to the LLM. In this case, the developer is responsible for deciding which trends are important and how to describe them in text for the LLM. Alternatively, we can visualize the simulation data and provide these visualizations as input to multi-modal LLMs (i.e., that can handle both text and images) such as OpenAI's GPT-4o or Google's Gemini. In this case, the developer uses several visualizations to reduce data while preserving spatial patterns (to know how agents interact with their environment) or variability (to express uncertainty). The visualizations can be advanced as they are intended for use by the LLM rather than by participants. Examples of visualizations evaluated with a panel of modelers are provided in~\citet{giabbanelli2019visual}.

After combining the LLM-generated text generated for the complete model's structure and insights from the simulations, we can summarize based on controllable aspects: what a stakeholder wants to know (content or information coverage) and how it should be expressed (writing style). At this stage, we do not yet know their preferences so we only generate different \textit{candidate} summaries to elicit feedback that will guide the next text generation. We thus recommend using designed experiments to generate several summaries based on content, style, and their interaction. For instance, if each controllable aspect is simplified by two options, then we have $2^2$ factorial design, resulting in four summaries.

Before evaluating the summaries across stakeholders, we need to ensure correctness. There is little value in producing summaries that speak about the right topics in the desired style but it may support the wrong decisions based on a misleading interpretation of the model. Since evaluating summaries for factuality with crowdsourced workers may not be reliable~\cite{zhao2023felm}, we recommend evaluating the summaries with modelers based on a questionnaire that covers three dimensions; whether each text contains factual errors (i.e., factuality labeling) and if so, explain the issues (error reasons) and categorize them (error types) as a knowledge error (hallucinated or inaccurate information), a reasoning error (flawed logic or reasoning), or irrelevant (content unrelated to the case). Each text should be evaluated by at least two modelers and their agreement rate can be calculated using weighted kappa to ensure reliability~\cite{tran2020weighted}. If the score is unsatisfactory then the error must be located (was it the summarization? the model-to-text? the simulation-to-text?), the generation process revised, and the summaries re-evaluated. 

Once the `candidate' summaries are technically correct, we can pilot them with stakeholders to elicit and measure reactions depending on the specifics of a project. In the context of health, \textit{empathy} is especially important as policy decisions impact vulnerable groups. Empathetic stories trigger emotional engagement, which motivates action-oriented decisions, thus supporting the translation of simulation outcomes into practice. Although there are several validated empathy questionnaires~\cite{lima2021empathy}, time-efficient options are particularly valuable to maximize participation and response quality~\cite{jeong2023exhaustive}. Since perceiving a narrative as immersive and compelling depends on the mental state of the reader, we recommend using the validated Toronto Empathy Questionnaire (TEQ; 16 items) to obtain multiple empathy measures~\cite{spreng2009toronto}. It has been used both for human readers who evaluate LLM-produced narratives~\cite{shen2024heart} and to test an LLM’s direct ability at producing empathetic text~\cite{welivita2024chatgpt,schaaff2023exploring}. For a comprehensive review on psychometric instruments applicable to LLMs, we refer the reader to~\citet{ye2025large}. In short, each participant would complete TEQ then receive four summaries (corresponding to the combinations of two controllable attributes) and complete a validated questionnaire for each one, such as the State Empathy Scale (12 items). We pilot-tested this protocol on measuring empathy from LLM-generated summaries with seven participants and noted a median response time of 19.16 minutes. The reasons behind participants' preferences and attitudes in the surveys can be further studied by one-on-one interviews.

\subsection{Step 2: Optimize the Alignment of Language Models and Stakeholder Communication Needs}

As the set of summaries were generated based on a design of experiments, a \textit{factorial analysis} can decompose the participants' reactions as a function of the controllable aspects (information content and writing style). This analysis should be performed for each group of stakeholders, as the goal is to generate insight into the preferred modalities of each group. Note that a `group of stakeholder' is not necessarily defined by role (e.g., patients, caregivers, physicians, healthcare administrator), so a complementary study may be needed to identify meaningful clusters~\cite{lavin2018should,aminpour2021social}. Additional analyses can include effect sizes to indicate the magnitude of the observed effects, a post-hoc power analysis to examine whether the sample size was adequate, Cronbach’s alpha to evaluate the internal consistency of each instrument within the specific population of respondents (as the questionnaire were validated in a more general sample), and repeated measures ANOVA to analyze how reactions vary across summaries. If qualitative one-on-one interviews were performed, then discourse analysis can examine how different groups use language, as the interviews may reveal differing narratives, framings, or ideological stances.

\begin{figure*}[!t]
\centering
\includegraphics[width=1\textwidth]{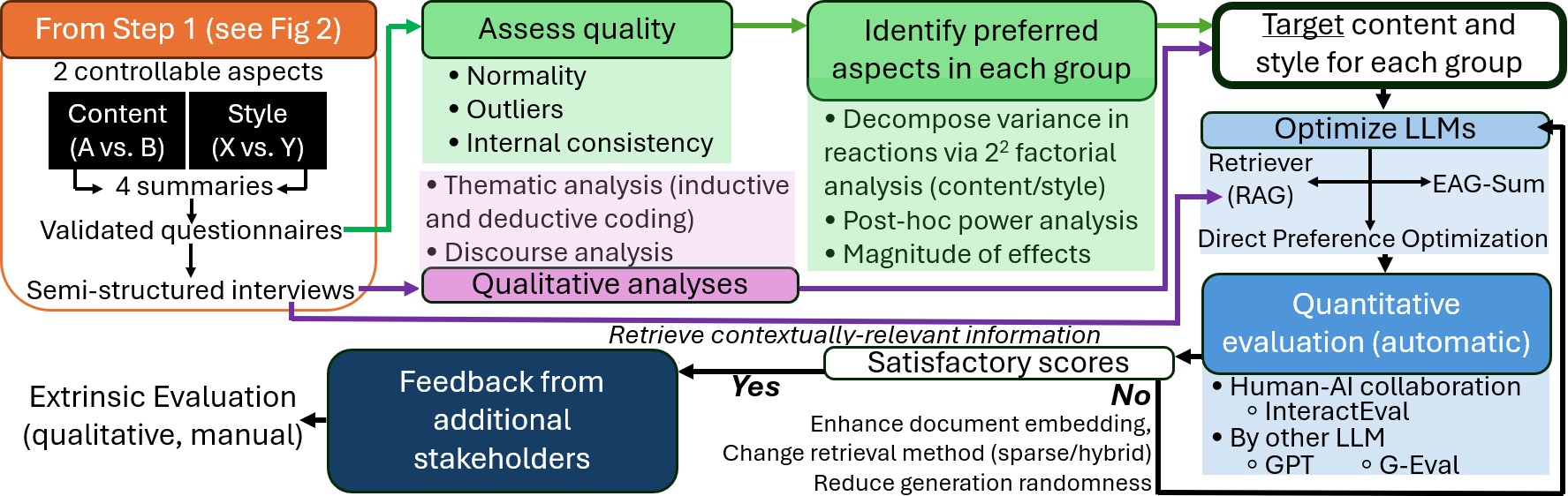} 
\caption{Our second step analyses the survey data to find what each group needs in a summary. Then, we steer LLMs in producing summaries that match these needs. An optimization process is involved, as the new summaries should be automatically assessed and the architecture adjusted if the scores are insufficient. Finally, the results can be presented to stakeholders.}
\label{fig:Step3}
\end{figure*}

After completing the data analysis to identify the preferred features, we steer the LLM to generate new summaries that match the preferred content and styles of each group. \textit{User-centric summarization} is challenging as people have long struggled to determine what is a good summary~\cite{zhao2009makes}. Useful summaries depend on three types of contextual factors: \textit{input factors} (the material that will be summarized), \textit{purpose factors} (the intended purpose of the summary), and \textit{output factors} (the characteristics of the generated summary).

Recent advances in text summarization encompass a spectrum of approaches (extractive, abstractive, generative, emerging hybrid architectures), each offering unique strengths. Extractive summarization selects salient sentences or phrases directly from the source text, maximizing factual retention and minimizing distortion, whereas abstractive methods rephrase and restructure ideas to produce more natural, human-like summaries. Generative models, particularly those built on transformer architectures, have emerged as powerful tools capable of synthesizing information across multiple documents and producing contextually rich narratives, though sometimes at the expense of factual precision.

\citet{ahmed2025hybrid} demonstrate that extractive models excel in factual accuracy, abstractive models achieve greater coherence and conciseness, and generative models capture subtle contextual nuances. Their proposed {\ttfamily EAG-Sum} hybrid framework strategically integrates all three paradigms in a multi-stage process: (1) generating an extractive ``skeleton'' summary with a transformer-based model such as {\ttfamily BERTSum} to ensure coverage of core factual details; (2) refining this skeleton with an abstractive model like {\ttfamily PEGASUS} or {\ttfamily BART} to improve fluency, reduce redundancy, and introduce novel sentence structures; and (3) applying generative contextual enhancement via a model such as GPT-series to adapt tone, style, and domain specificity, adding auxiliary detail where needed. 

There remains significant scope for improvement in steering LLMs toward accurate and desirable outputs through domain adaptation and human feedback-driven alignment strategies. Traditional supervised fine-tuning (SFT) alone is insufficient, as it optimizes against static gold-standard labels and fails to capture the richness of human preferences in open-ended generation. Recent approaches instead treat alignment as a preference optimization problem, where models are adapted to approximate the behavior preferred by human evaluators. Reinforcement Learning from Human Feedback (RLHF) has become a standard paradigm: a reward model is trained on human preference comparisons, and the base language model is then fine-tuned via reinforcement learning to maximize expected reward. Direct Preference Optimization (DPO)~\cite{10.5555/3666122.3668460} reformulates alignment by removing the explicit reward model. Instead, it optimizes the language model parameters to directly maximize the log-likelihood ratio of preferred vs. dispreferred responses, using pairwise preference data. RAFT (Reward-ranked Fine-Tuning)~\cite{dong2023raft} and RLAIF (Reinforcement Learning from AI Feedback)~\cite{lee2023rlaif} extend these ideas by reducing the dependence on costly human labels. Diverse AI Feedback~\cite{yu2025diverse} goes beyond pairwise preferences by incorporating heterogeneous forms of supervisory signals into the optimization objective. Specifically, it integrates: (1) critique feedback, which provides structured error annotations and diagnostic signals; (2) refinement feedback, in which annotators (or auxiliary models) propose partial rewrites at the span or sentence level; and (3) ranking-based preferences, which offer global desirability signals across candidate completions. Using proven and integrated solutions can simplify the process instead of manually combining some of the (slightly) older tools such as a generation that handles multiple levels of details like {\ttfamily GranuSum}~\cite{zhong2022unsupervised}, a method to identify relevant facts for a target audience~\cite{hayashi2021wikiasp}, and yet another approach to control stylistic variations in tone or readability~\cite{ribeiro2023generating}. 

While the approaches mentioned above provide a comprehensive set of tools to generate summaries, there is no guarantee that they will be optimal at first. For example, recent research has exposed counterintuitive pitfalls: \citet{peters2025generalizationbiaslargelanguage} show that explicitly instructing LLMs to produce more faithful summaries can backfire, increasing overgeneralization rates by up to 15\% in some models. Assessment is thus needed and may be followed by mitigation strategies such as changing some of the parameters of the RAG (Figure~\ref{fig:Step3}) or reducing generation randomness (e.g., lowering temperature to 0). The assessment should ensure that the newly generated summaries continue to logically follow from the content that was approved by modelers for factuality. While methodological advances have expanded the capabilities of summarization systems, their evaluation remains a critical and evolving challenge. Evaluating a large number of candidate summaries with respect to the preferences of each group would become time-consuming for human readers and it may not be mindful of the time commitment of participants. Thus, LLMs can be used to perform some of the evaluation. Studies have shown that GPT produced better preference and factuality ratings than conventional evaluation metrics on several datasets~\cite{gao2023human}. Multi-dimensional evaluation metrics such as UniEval \cite{zhong2022towards} correlate more strongly with human judgments \cite{fraile2025expert}, while another LLM-based method, G-Eval~\cite{liu2023g}, demonstrated a Spearman correlation coefficient above 0.5 with human Likert-scale judgments on news summaries using GPT-4. More recently, innovative \textit{human–AI collaborative evaluation} approaches, such as InteractEval~\cite{chu2025think}, combine the high-level reasoning and flexibility of human Think-Aloud protocols with the consistency and breadth of LLM-generated checklists, producing superior benchmark performance on summarization datasets. The analysis by~\citet{chu2025think} reveals that humans excel at identifying internal quality attributes (coherence, fluency), while LLMs better capture external alignment (consistency, relevance), suggesting the value of integrating both perspectives.


Once optimized summaries have been produced for each stakeholder group, the next step is to share them back with participants. This step must be approached with the understanding that stakeholders are not simply a source of data for iterative model refinement. They are often busy professionals or community members who volunteer their time because they care about solving a real problem. If we ask them to engage again to provide feedback on improved LLM-generated summaries, the interaction should deliver tangible value to them as well. We therefore recommend presenting the results in formats that benefit participants, such as workshops or educational sessions, where they can both learn from the findings and connect with other stakeholders.

\section{Discussion: On Participatory AI}

Given the importance of engaging different groups of users with modeling and simulation, we articulated a vision and complete process that leverages advances in (multimodal) LLMs to produce summaries tailored to the informational needs and styles of each group. Our framework is intentionally broad and mixed-methods, as we need several measures to capture the needs of participants and the extent to which a summary addresses these needs. The scientific basis for this framework will be strengthened by collecting experimental data, performing ablation studies to measure the effect of each part of the framework (which may result in a simplified framework), and comparing strategies (e.g., few-shot learning vs. supervised fine-tuning vs. preference optimization). While our vision focused on measures related to content and style, there are also \textit{operational metrics}. If modelers visit low-resource settings where connectivity and hardware are constrained, they still need the LLM to generate text in a timely manner (e.g., measure response time). This may call for architectures centered on lightweight open-source LLMs and/or hybrid pipelines with edge computing to push queries on a user's hardware (to the extent possible) and only depend on the cloud for occasional secure retrieval. 

In practice, stakeholders rarely read a summary, instantly trust its conclusions, and know exactly how to act. They may question the assumptions behind the model, the impact of alternative scenarios, or the reasons for specific simulation outputs. Provenance information is critical for answering these questions~\cite{gierend2024provenance}, especially when results are counter-intuitive or central to justifying an intervention. In our context, however, provenance is complex: generated summaries can vary with LLM parameters (e.g., temperature, stochastic routing in mixture-of-experts models), the order of graph linearization, or the retrieval index used. This leads to a human-factors challenge: we are asking participants to trust explanations generated by an LLM (a technology they may already distrust) in order to build trust in a simulation model, which may also mistrusted. As ~\citet{hinrichs2025ai} note in the context of mental health, ``barriers to the integration of AI primarily stem from issues related to trust and confidence in the system, end-user acceptance, and system transparency''. While this trust paradox is real, the alternative is less desirable, since continuing with the current status quo would mean leaving participants to face significant barriers while engaging with modeling.

Going forward, we thus envision the creation of more interactive environments that \textit{extend} the textual summaries discussed in this paper. This immediately raises the question: \textit{how} would individuals interact with the content? For example, the field of \textit{visual analytics} has long been applied to healthcare and it supports interactions through details-on-demand on linked visualizations~\cite{caban2015visual}. In our prototype, a part of the summary can be expanded into complete paragraphs and the corresponding part of the model is provided as a node-and-link diagram~\cite{gandee2024visual}. However, the need to create text in this paper was motivated by the difficulty of engaging various audiences with scientific visualizations, thus embedding the text within overly technical platforms may defeat this purpose. An alternative could be to promote simpler \textit{conversational} interactions by voice or text, but they raise numerous technical challenges to clarify a user's question and provide answers by combining information~\cite{giabbanelli2024broadening}.

Our framework articulated numerous qualitative and quantitative assessments to ensure that each group is presented with a summary that is factual and addresses their preferences. But we should not lose track of the bigger picture: the text is not the end goal for assessment. We create summaries to support decision-making activities in each group, so the ultimate demonstration that the pipeline works lies in its ability to affect decisions. We thus need to broaden evaluation to include downstream decision metrics: does a tailored summary change the decisions stakeholders make compared to a generic summary? A randomized controlled trial may assign stakeholders to different summaries (e.g., generic vs. tailored) to measure the impact of the decision. 


\end{document}